\documentclass[10pt,twocolumn,letterpaper]{article}

\usepackage{iccv}
\usepackage{times}
\usepackage{epsfig}
\usepackage{graphicx}
\usepackage{amsmath}
\usepackage{amssymb}

\usepackage[pagebackref=true,breaklinks=true,letterpaper=true,colorlinks,bookmarks=false]{hyperref}

\iccvfinalcopy

\ificcvfinal\fi

\usepackage{soul}
\usepackage{url}
\usepackage[utf8]{inputenc}
\usepackage[small]{caption}
\usepackage{booktabs}

\usepackage{amssymb}
\newcommand{\R}{\mathbb{R}}
\newcommand{\norm}[1]{\left\lVert#1\right\rVert}

\usepackage{subcaption}
\usepackage{tabularx}
\usepackage{multirow}
\newcolumntype{Y}{>{\centering\arraybackslash}X}

\usepackage{color, colortbl}
\definecolor{Gray}{gray}{0.9}
\begin{document}

\title{Cascading Convolutional Temporal Colour Constancy}

\author{
    \textbf{Matteo Rizzo}, \textbf{Cristina Conati}\\
    University of British Columbia\\
    {\tt\small \{marizzo, conati\}@cs.ubc.ca}
    \and
    \textbf{Daesik Jang}, \textbf{Hui Hu}\\
    Huawei\\
    {\tt\small \{daesik.jang, huhui12\}@huawei.com}
}

\maketitle
\ificcvfinal\thispagestyle{empty}\fi

\begin{abstract}
\noindent Computational Colour Constancy (CCC) consists of estimating the colour of one or more illuminants in a scene and using them to remove unwanted chromatic distortions. Much research has focused on illuminant estimation for CCC on single images, with few attempts of leveraging the temporal information intrinsic in sequences of correlated images (e.g., the frames in a video), a task known as Temporal Colour Constancy (TCC). The state-of-the-art for TCC is TCCNet, a deep-learning architecture that uses a ConvLSTM for aggregating the encodings produced by CNN submodules for each image in a sequence. We extend this architecture with different models obtained by (i) substituting the TCCNet submodules with C4, the state-of-the-art method for CCC targeting images; (ii) adding a cascading strategy to perform an iterative improvement of the estimate of the illuminant. We tested our models on the recently released TCC benchmark and achieved results that surpass the state-of-the-art. Analyzing the impact of the number of frames involved in illuminant estimation on performance, we show that it is possible to reduce inference time by training the models on few selected frames from the sequences while retaining comparable accuracy.
\end{abstract}

\section{Introduction}
\label{sec:introduction}

\noindent From the perspective of human cognition, Colour Constancy (CC) is the ability of the visual system to perceive colours of objects invariant of the colour cast generated by the lighting conditions of a scene \cite{color-constancy}. Computationally, CC (also known as Computational Colour Constancy, or CCC) consists of estimating the colour of one or more illuminants in a scene and using them to remove unwanted chromatic distortions by colour-correcting the input. CCC has attracted considerable attention in computer vision enabling the compensation of the effects of changing illumination in images and videos. Prominent applications of CCC relate to off-line tasks such as the preprocessing of these digital contents in datasets used to train models for computer vision tasks benefiting from intrinsic colour information (e.g., fine-grained object recognition and semantic segmentation), in domains such as video surveillance \cite{video-surveillance}, screening tools for healthcare \cite{screening-tools}, digital pathology \cite{digital-pathology}, etc., and the postprocessing of consumer photography. Compensation of changing illumination is also a key step of the processing pipeline of images and videos in digital cameras known as white balancing \cite{color-pipeline}. In this context, CCC is used in real-time for processing the raw data captured by the camera sensor, and allows, for instance, to constantly adjust the stream of frames displayed to the user in the camera viewfinder for enhanced chromatic fidelity.

\begin{figure}[t]
    \centering
    \begin{subfigure}{.49\textwidth}
         \includegraphics[width=\textwidth]{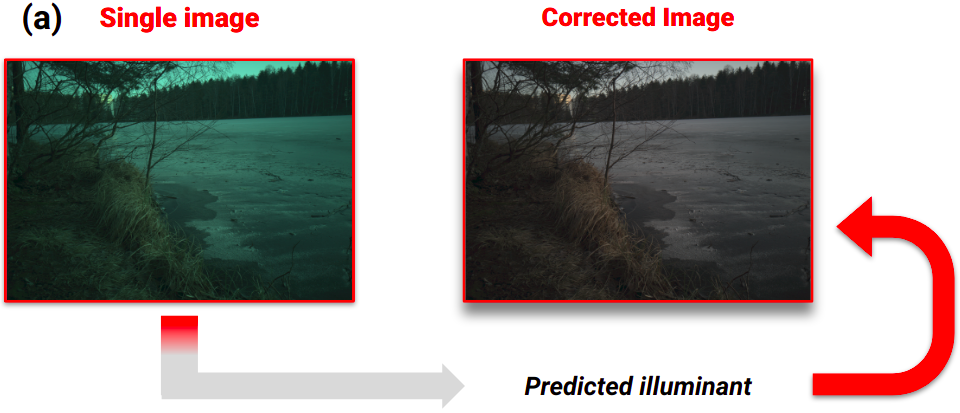}
         \label{fig:single-frame-task}
    \end{subfigure}
    \begin{subfigure}{.5\textwidth}
         \includegraphics[width=\textwidth]{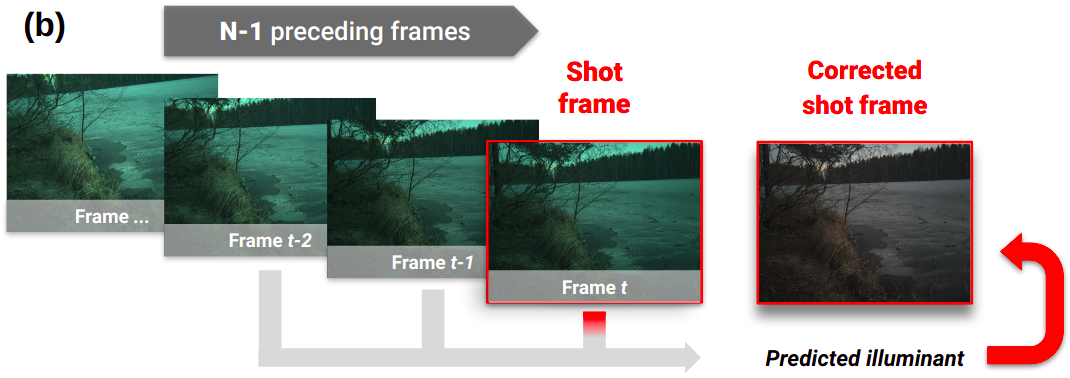}
         \label{fig:multi-frame-task}
    \end{subfigure}
    \caption{The single-frame CCC task (a) and the multi-frame, or temporal, CCC task (b).}
    \label{fig:ccc-tasks}
\end{figure}

Figure \ref{fig:ccc-tasks} shows CCC leveraging a single frame (i.e., an image) vs a sequence of frames (as in a video). In single-frame CCC, the illuminant for an input frame is estimated using only the information stemming from that frame. The multi-frame, or temporal, CCC task (TCC from now on) implies leveraging information from a window of frames preceding the one in the sequence for which the illuminant is being estimated (i.e., the so-called \textit{shot frame}).

Addressing the TCC task to improve video quality is increasingly important, as videos are becoming a predominant medium in all aspects of everyday life, especially after the surge of online video-based activities caused by the COVID-19 pandemic. However, TCC entails unique challenges such as changes of illuminant chromaticity and varying viewing angles of the camera \cite{rcc}. Note that, when dealing with sequences of images, a naive approach may consist of estimating an illuminant for each image individually. However, such a strategy does not account for the temporal information inherent in a sequence of correlated items (e.g., the frames of a video), which may be determinant in a realistic scenario featuring drastic changes of light sources and content in subsequent images. In this paper, we contribute to improving the state-of-the-art accuracy of TCC by extending an existing method that leverages the temporal cues intrinsic to sequences of correlated images.

\begin{figure*}[t]
    \centering
     \includegraphics[width=.8\textwidth]{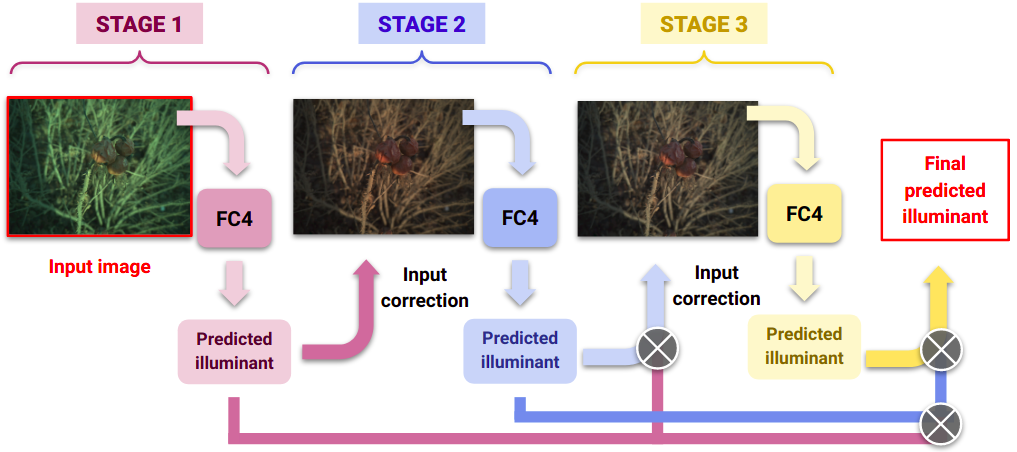}
    \caption{The cascading strategy implemented by the C4 method.}
    \label{fig:c4}
\end{figure*}

The TCC task was formally introduced in \cite{rcc} as a general illuminant estimation for a sequence of images, relaxing strong assumptions such as constant illuminant across frames and fixed-length sequences. More specifically, given a sequence of frames $I_i$ of arbitrary length, the illuminant colour of the $t$-th frame $\hat{c_t} = [\hat{y_r},~\hat{y_g},~\hat{y_b}] \in \R^3$ (assumed to be only one) is estimated by some function $f(\cdot)$ considering the $N-1$ (with $1 \le N \le t$) frames preceding it (Formula \ref{eq:tcc-task}). Note that single-frame CCC is a special case of the temporal task where $N=1$ (i.e., only the shot frame is considered). 

\begin{equation}
    \label{eq:tcc-task}
    \hat{c_t} = f(I_{t-(N-1)}, I_{t-(N-2)} ... I_{t-1}, I_t)
\end{equation}

Most of the research so far has focused on CCC for single frames (see \cite{ccc-survey} for a survey), with some attempts of leveraging the temporal information intrinsic in sequences of correlated frames for illuminant estimation \cite{rcc,tcc}. In both cases, state-of-the-art methods in terms of accuracy rely on Deep Learning (DL).

In this paper, we investigate how to advance the state-of-the-art in TCC by proposing three new architectures that integrate the current  best-performing model for TCC \cite{tcc} with a strategy that proved successful in the single-frame scenario \cite{c4} but which is novel to the temporal task. Our results show that the novel approaches surpass in terms of accuracy the state-of-the-art on the TCC benchmark \cite{tcc}, which is currently the only dataset specifically designed for experimenting with CCC in the temporal scenario. We also present a detailed analysis of the trade-off between model accuracy and computational cost (in terms of inference time and memory occupancy), relevant for the practical applications of CCC. As part of this analysis, we investigate the number of frames involved in the illuminant estimation as a possible way to reduce the inference time of the TCC models. Contrary to preliminary results in \cite{tcc}, we show that it is possible to train models that work with shorter sequences, with substantial gain in inference efficiency and no major loss in accuracy, if the retained frames are selected to capture global temporal information. 

\section{Related Work}
\label{sec:related-work}

\noindent In the last decades there has been extensive research on  single-frame CCC, categorized by \cite{ccc-survey} in three groups of methods: (i) \textit{static methods}, (ii) \textit{gamut-based methods}, and (iii) \textit{learning-based methods}. Static methods operate on the basis of predefined statistical or physical properties of the image formation, whereas learning-based methods are capable of learning a mapping between some features of the input images and the target illuminants. Many gamut-based methods (such as \cite{gamut}) use training data to define a target gamut and can therefore be considered as learning-based methods. Recent DL-based methods such as \cite{ffcc,quasi-unsupervised-cc,cc-reweighting,c4} have proved the most accurate among learning-based methods and outperformed static methods with a notable discrepancy in performance. Among these DL methods for single-frame CCC, the state-of-the-art is represented by FFCC \cite{ffcc} and C4 \cite{c4}, performing similarly on standard benchmark datasets (i.e., NUS 8-Camera \cite{nus-dataset} and Color Checker \cite{color-checker-paper}). FFCC relies on quite an unusual yet very effective strategy that solves illuminant estimation by reducing it to a spatial localization task on a torus and operating in the frequency domain. The best performing FFCC model leverages a Convolutional Neural Network (CNN) in the optimization step. C4 tackles the problem with a more standard neural approach. This method builds on FC\textsuperscript{4} \cite{fc4}, a fully convolutional network architecture based on a pretrained backbone (i.e., either SqueezeNet \cite{squeezenet} or AlexNet \cite{alexnet}). C4 iterates FC\textsuperscript{4} in three stages while performing incremental corrections of the input image based on the illuminant estimate outputted at each stage. For developing some of the novel architectures we propose in this paper for the multi-frame scenario, we used the same cascading strategy introduced by C4 for the single-frame scenario (described in more detail in Section \ref{sec:tccnet}).

In contrast with  the plethora of methods for CCC targeting single images, few attempts have been made to deal with sequences of frames (i.e., with the temporal task). Research on TCC so far has either focused on some special cases (e.g., \cite{ac-bulb} target AC bulb illuminated scenes and \cite{flash} pairs of images with and without flash), or made strong assumptions such as constant illumination across frames \cite{video-based-estimations,video-sequences}. There have been only a few attempts that we are aware of devising more general-purpose neural strategies \cite{ffcc,rcc,tcc}. Our work fits the latter category. The previously mentioned FFCC method natively supports a multi-frame extension for temporal smoothing, but does not achieve state-of-the-art results \cite{ffcc}. \cite{rcc} propose a neural network architecture for illuminant estimation on multiple frames called RCCNet. This is based on encoding each of the single images in a sequence with an AlexNet CNN \cite{alexnet}, and then processing these encodings temporally with an LSTM recurrent network. RCCNet was shown to outperform other straightforward extensions of learning-based single-frame algorithms for processing multiple frames. The baseline established by RCCNet was recently overcome by TCCNet \cite{tcc}, an improved version of the same method featuring a more powerful backbone CNN and a ConvLSTM for the sequential processing of the encoded frames. We will describe TCCNet in more detail in later sections as this is the architecture we extended in our approach.

\section{Proposed Approaches}
\label{sec:proposed-approaches}

\noindent This section describes how we extended existing state-of-the-art CCC strategies to build our novel architectures. We first provide an overview of both C4 and TCCNet, the existing models that we leveraged, which target the single-frame and multi-frame scenarios respectively. Then, we detail the three different architectures we propose to integrate and improve such models.

\subsection{The C4 and TCCNet  Architectures}
\label{sec:c4-tccnet}

\subsubsection{C4 Architecture} 
\label{sec:c4}

\noindent In the C4 method, the illuminant estimate for an input image is built via a cascading strategy, illustrated  in Figure \ref{fig:c4}. In each of the $L=3$ stages of the cascade, a SqueezeNet \cite{squeezenet} backbone produces a prediction of the colour of the illuminant for its input. This prediction is then used to generate a colour-corrected version of the original image, which is fed to the next stage in the cascade. Formally, given an image $I$ and a ground truth colour of the illuminant $c_{gt}$, the estimate $\hat{c} = [\hat{y_r},~\hat{y_g},~\hat{y_b}] \in \R^3$ of $c_{gt}$ is used to generate the corrected image $\bar{I} \in \R^{H \times W \times 3}$ so that the original input results under some standard illuminant, usually white light (Formula \ref{eq:color-correction}). The correction assumes that each RGB channel can be adjusted independently \cite{chromatic-adaption}.

\begin{equation}
    \label{eq:color-correction}
    \bar{I_j} = I_j / \hat{c_j} \in \rm I\!R^{H \times W},~j \in \{r, g, b\} 
\end{equation}

The colour $\hat{c_i}$ used for correcting the input at stage $i$ is obtained by multiplying the current estimate with those produced in the previous stages (Formula \ref{eq:color-estimate}).

\begin{equation}
    \label{eq:color-estimate}
    \hat{c_i} = \prod_{k=1}^{i} c_k,~i \in [1, L]
\end{equation}

\subsubsection{TCCNet Architecture} 
\label{sec:tccnet}

\begin{figure}[t]
    \centering
     \includegraphics[width=.5\textwidth]{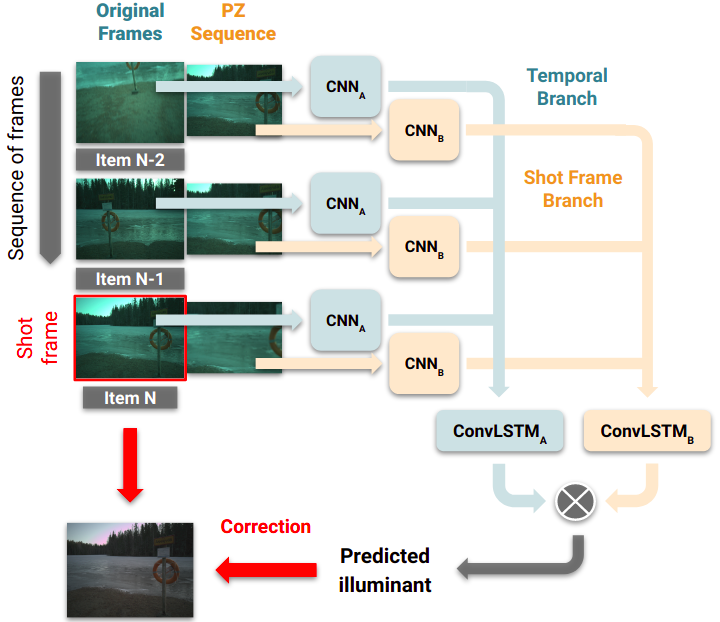}
    \caption{The TCCNet architecture. Our proposed TCCNet-C4 architecture substitutes the CNN submodules with C4.}
    \label{fig:tccnet}
\end{figure}

\noindent The architecture of TCCNet is described in Figure \ref{fig:tccnet}. It consists of two branches: (i) a \textit{temporal branch} processing the original sequence of images and (ii) a \textit{shot frame branch} processing a Pseudo-Zoom (PZ) sequence in the shot frame. The PZ sequence strategy for the TCC task was introduced in \cite{rcc} and consists of multiple frames generated on a random zoom-path (i.e., a series of sub-patches based on consecutive geometric transformations) for a target image. Both branches follow the same CNN + LSTM structure and their output is merged for producing the final classification. In the original architecture, the CNN submodules are instantiated to pretrained SqueezeNet \cite{squeezenet} models. The LSTM network is a 2D ConvLSTM using convolutional structures in both input-to-state and state-to-state transitions, as proposed by \cite{conv-lstm}. The original frames and the PZ sequence are passed to the CNN submodules to generate two parallel sequences of encodings. These are then processed by as many ConvLSTMs, whose output is concatenated to generate the estimated illuminant for the shot frame.

\begin{figure*}[t]
    \centering
     \includegraphics[width=.75\textwidth]{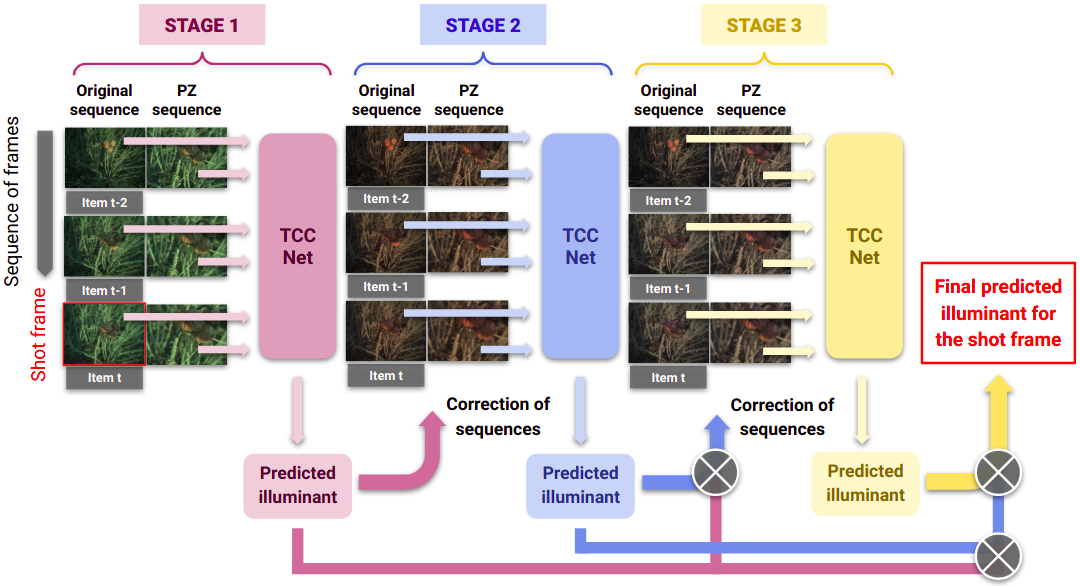}
    \caption{Our proposed C-TCCNet architecture. In C-TCCNet-C4 we replaced the TCCNet submodules with TCCNet-C4.}
    \label{fig:ctccnet}
\end{figure*}

\subsection{Novel Architectures}
\label{sec:novel-architectures}

\noindent In an attempt to improve the performance of the best existing neural methods for TCC, we propose three novel architectures that leverage either C4, TCCNet or both: (i) \textit{TCCNet-C4}, (ii) \textit{Cascading TCCNet} (C-TCCNet) and  (iii) \textit{C-TCCNet-C4}, described in the rest of this section.

\subsubsection{TCCNet-C4 Architecture}
\label{sec:tccnetc4}

Whereas the CNN submodules in the branches of the original TCCNet architecture (see green and orange squares labelled as $CNN_A$ and $CNN_B$ at the center of Figure \ref{fig:tccnet}) are substantially SqueezeNet, in TCCNet-C4 we replaced them with instances of C4. The final linear layer on top of the last stage in the cascade used by the C4 submodules is removed to feed the generated encodings directly to the recurrent component. The rationale for substituting C4 submodules to the original SqueezeNets  is to provide the TCCNet architecture with a  more powerful backbone to encode each frame in a sequence, given that C4 is the state-of-the-art neural architecture for single-frame CCC. This is achieved by applying the cascading colour correction to each image in the sequence independently, which intuitively results in feeding to the temporal component a series of encodings of frames whose colour of the illuminant is more resembling the ground truth (and thus easier to process). 

\subsubsection{C-TCCNet and C-TCCNet-C4 Architectures}
\label{sec:ctccnet-ctccnetc4}

\noindent Inspired by the cascading approach that C4  proved to be successful in the single-frame CCC scenario, we sought to apply it to the TCC task. Following a similar strategy to that implemented by the  C4 method, we plugged the TCCNet and TCCNet-C4 models in a cascading architecture to build the C-TCCNet and C-TCCNet-C4 architectures respectively. The network architecture for the C-TCCNet models is presented in Figure \ref{fig:ctccnet}. These novel strategies allow for iterative coarse-to-fine refinement of the estimate of the illuminant of the shot frame accounting for information stemming from the temporal sequence. Namely, each stage in the cascade outputs a prediction for the illuminant of the shot frame, which is used to generate the input to the following stage by colour-correcting each frame in both the original sequence and PZ sequence. In the same fashion as C4, the estimated colour used for the cascading correction is the product of the current estimate by the predictions at the previous stages. By incorporating C4 as a submodule, the C-TCCNet-C4 architecture leverages a cascading-correction approach both in the per-frame and sequence-wise estimate.  

\subsection{Implementation}
\label{sec:implementation}

\noindent All our presented models were built and trained using the PyTorch\footnote{https://pytorch.org/} library. The code is available in a dedicated open-source repository\footnote{Link omitted for preserving anonymity} which, aside from the novel TCCNet-C4, C-TCCNet and C-TCCNet-C4 architectures, also features a clean re-implementation of the baseline TCCNet and C4 (for better integration with the rest of our codebase).  

\section{Models Evaluation}
\label{sec:evaluation}

\subsection{Dataset and Settings for Model Training}
\label{sec:dataset}

\noindent We evaluated our models on the TCC dataset \cite{tcc}, which is the largest and most realistic dataset available for TCC\footnote{Of the other existing datasets for TCC, \cite{video-sequences} is  very small, \cite{ac-bulb} was specifically designed for experiments on AC bulb illumination and \cite{sfu-grayball} features very low-resolution images which are not in line with the standards of modern consumer photography.}. It consists of 600 real-world videos recorded with a high-resolution mobile phone camera shooting 1824 x 1368 sized pictures. The length of these videos ranges from 3 to 17 frames (7.3 on average, the median is 7.0 and mode is 8.5). Ground truth information is present only for the last frame in each video (i.e., the shot frame), and was collected  using  a gray surface calibration target.

Despite TCC being the largest dataset available, the number of sequences is relatively low for training a DL model effectively. Data augmentation is therefore performed according to the same procedure used in \cite{tcc} to train TCCNet. Namely,  the frames in the sequences were randomly rotated by an amount in the range [-30, +30] degrees and cropped to a proportion in the range of [0.8, 1.0] on the shorter dimension. The generated patches were flipped horizontally with a probability of 0.5. Data augmentation is achieved by dynamically applying the aforementioned transformations at training time to each batch of sequences, which is the standard practice in PyTorch. 

All models have been trained using a mix of  Tesla P100 and NVidia GeForce GTX 1080 Ti GPUs from local lab equipment and cloud services (name withdrawn to presence anonymity) and took about two to four days based on the complexity of the model. The training of the TCCNet models was performed for $2k$ epochs using the RMSprop \cite{rmsprop} optimizer with batch size 1 and learning rate initially set to $3e^{-5}$. We opted for hidden size = 128 and kernel size = 5, as suggested by the ablation study in \cite{tcc}. The SqueezeNet backbone was initialized with the weights pretrained on ImageNet \cite{imagenet} provided by PyTorch\footnote{The pretrained models offered by PyTorch are available at https://pytorch.org/docs/stable/torchvision/models.html}. The error $\epsilon$ between the illuminant $\hat{c}$ estimated by the TCCNet models and the ground truth $c_{gt}$ has been computed using the angular error, a measure of error used in many works on CCC and reported in Formula \ref{eq:angular-error}.

\begin{equation}
    \epsilon_{\hat c, c_{gt}} = arccos(\frac{\hat c \cdot c_{gt}}{\norm{\hat c} \norm{c_{gt}}})
    \label{eq:angular-error}
\end{equation}

Using the same optimizer, batch size and learning rate, the cascading models were trained for $1k$ epochs initializing the submodules to the corresponding pretrained TCCNet models. $L=3$ was selected as the number of stages for each cascade as an analysis of the length of the iteration in \cite{c4} showed how this is a good trade-off between performance and complexity of the model. The measure of the error of choice for these models was the multiply-accumulate loss function $\mathcal{L}$ for cascading convolutional CCC introduced in \cite{c4} and reported in Formula \ref{eq:mal}. 

\begin{equation}
    \mathcal{L} = \sum_{l=1}^{L} \mathcal{L}^{(l)} (\prod_{i=1}^{l} f_i (X_i), y)
    \label{eq:mal}
\end{equation}

This loss function is meant to enforce coarse-to-fine optimization of each of the stages in the cascade via supervision of the intermediate estimates. The intermediate error scored at the $l$-th stage of the cascade with respect to the input $X$ of ground truth $y$ is monitored using the angular error loss (denoted as $\mathcal{L}^{(l)}$). This is computed based on the product of the current estimate $f(X)$ with all those preceding it.

\begin{table*}[t]
    \centering
    \small
    \begin{tabular}{lcccccc}
    \toprule
    Model & Mean & Median & Trimean & Best 25\% & Worst 25\% & Worst 5\% \\
    \midrule
    TCCNet & 1.99 & 1.21 & 1.46 & 0.30 & 4.84 & 6.34 \\
    C-TCCNet & 1.95 & 1.22 & 1.42 & 0.25 & 4.78 & 6.10 \\
    TCCNet-C4 & \underline{1.72} & \underline{1.08} & \underline{1.20} & \textbf{0.20} & \textbf{4.33} & \textbf{6.01} \\
    C-TCCNet-C4 & \textbf{1.70} & \textbf{0.92} & \textbf{1.13} & \underline{0.22} & \underline{4.36} & \underline{6.05} \\
    \bottomrule
    \end{tabular}
    \caption{Distribution of the angular error for the proposed methods and TCCNet on the TCC benchmark split (best results are \textbf{bolded}, second best are \underline{underlined}).}
    \label{tab:tcc-split}
\end{table*}

\begin{table*}[t]
    \centering
    \small
    \begin{tabular}{lcccccccccccc}
    \toprule
    Model & Mean & Median & Trimean & Best 25\% & Worst 25\% & Worst 5\% \\
    \midrule
    TCCNet & 1.98 (0.18) & 1.17 (0.14) & 1.38 (0.14) & 0.29 (0.02) & 4.98 (0.50) & 6.18 (0.22) \\
    C-TCCNet & 1.94 (0.18) & \underline{1.14} (0.12) & 1.34 (0.10) & \underline{0.25} (0.02) & 4.91 (0.59) & \underline{5.88} (0.66) \\
    TCCNet-C4 & \underline{1.86} (0.22) & \underline{1.14} (0.19) & \underline{1.32} (0.22) & 0.26 (0.07) & \underline{4.66} (0.45) & 6.00 (0.61) \\
    C-TCCNet-C4 & \textbf{1.79} (0.23) & \textbf{1.10} (0.26) & \textbf{1.25} (0.27) & \textbf{0.22} (0.04) & \textbf{4.55} (0.48) & \textbf{5.71} (0.86) \\
    \bottomrule
    \end{tabular}
    \caption{Distribution of the angular error for the proposed methods and TCCNet over 4 training-test splits from the TCC dataset. All metrics are reported in terms of average and standard deviation (best results are \textbf{bolded}, second best are \underline{underlined}).}
    \label{tab:4-splits}
\end{table*}

\subsection{Accuracy Metrics}
\label{sec:metrics}

\noindent As in previous works on CCC, the metrics we selected to evaluate model performance in terms of accuracy provide insights into the distribution of the angular errors across the test items. These metrics include the Mean Angular Error (MAE), the median and the trimean (defined as the weighted average of the median and upper and lower quartiles) across the test set, indicating how the models performed on average and accounting for outliers. We also report the MEA on the best 25th and worst 25th and 5th percentiles, showing how the model performed on easy, hard, and very hard inputs, respectively.

\subsection{Results}
\label{sec:results}

\noindent In this section we first present results related to the performance of our proposed models when using the full sequences from the TCC dataset, namely performance in predicting the illuminant of the last frame in each sequence (the only one for which the ground truth is available) using all the preceding frames. Performance is discussed  both in terms of accuracy and efficiency. Then, we present results related to the impact on performance of reducing the number of frames used for illuminant estimation (\textit{estimating frames} from now on).

\subsubsection{Accuracy on full sequences}
\label{sec:accuracy-full-sequences}

\noindent In this section we evaluate models accuracy in estimating  an illuminant for the last frame in a sequence (i.e., the shot frame) by leveraging all the preceding frames.

The TCC dataset comes with a fixed training-test split (with 2/3 of the data used for training, and 1/3 for testing) for benchmarking, which is what was used to evaluate the original TCCNet approach in \cite{tcc}. For a fair comparison with results previously achieved on the same benchmark, we first trained and tested the models (our three proposed models as well as our re-implementation of TCCNet) on the split provided with the dataset. These results are presented in Table \ref{tab:tcc-split}. For this data split, all our proposed models outperform the original TCCNet on all reported metrics (the only exception is the median error for C-TCCNet, which is slightly worse). Among our models, those which rely on C4 (C-TCCNet-C4 and TCCNet-C4)  show lower errors with respect to the one that does not  (C-TCCNet). In particular, C-TCCNet-C4 is the best performing model in terms of measures of central tendency, while TCCNet-C4 achieves slightly better scores on the easiest and hardest samples. 

In addition to the above comparison on the original data split in \cite{tcc}, we also tested the models on three alternative 2/3-1/3 splits, to achieve more robust results beyond the evaluation on a single split. Table \ref{tab:4-splits} reports model performance over the 4 splits tested\footnote{The three additional splits have disjoint test sets as in standard cross-validation. The test set of the original spit has approximately a 33\% overlap with the others, on average.  We use this approach rather than a full 4-fold cross-validation because of  the amount of time needed to train the proposed models  on one data split (between two and four days depending on the model).}, in terms of average and standard deviation of the performance metrics. In terms of sheer numbers, results are consistent with the benchmark split as all the novel models we propose show better values  than TCCNet on all the evaluation metrics. C-TCCNet-C4, in particular, shows the lowest values on all the metrics. TCCNet-C4 performs better than C-TCCNet on the measures of central tendency, but slightly worse on some of the extreme samples. 

We performed a statistical comparison of the performance of each of our proposed models against TCCNet, focusing on the mean error (first column in Table \ref{tab:4-splits}) as a general measure of central tendency that we used to formally rank our models. The analysis is done  via one-tailed paired t-tests with the Benjamini-Hochberg adjustment \cite{benjamini-hochberg} to account for multiple comparisons (3 in total) and effect sizes (a well-recognized  indication of practical significance) measured via Cohen’s D \cite{effect-size}. This statistical analysis shows significant improvement, with a large effect size\footnote{Following \cite{effect-size}, we consider the effect size large if $d \ge 0.8$, medium if $d \ge 0.6$, small otherwise.} for C-TCCNet-C4 $(p<0.05,~d=0.9050)$, as well as a significant improvement with a small effect size $(p<0.05,~d=0.2380)$ for C-TCCNet. Such findings suggest that it is the strategy of cascading TCCNet architectures  the aspect of our proposed approaches  that has the more solid contribution in achieving a lower mean error of the predicted colour of the illuminants.

\subsubsection{Efficiency on full sequences}
\label{sec:efficiency-full-sequences}

\noindent In light of the multiple applications of colour constancy, ranging from off-line tasks to real-time processing and involving deployment on both low-end and high-end hardware, in this section we investigate the trade-off between model accuracy and time/space efficiency. In particular, we focus on the baseline TCCNet model and our two proposed models that generated the best performance in the previous section, namely C-TCCNet and C-TCCNet-C4.

Table \ref{tab:resources-impact-seq-len} compares the three aforementioned models in terms of accuracy, inference time, and model size. Whereas accuracy is generally the primary concern for off-line tasks, inference time is particularly relevant for real-time applications, such as processing the raw data captured by the sensor of a digital camera, while model size has implications in the deployment on low-memory hardware (e.g., commodity mobile devices). In Table \ref{tab:resources-impact-seq-len}, accuracy is reported as the Mean Angular Error (MAE) averaged across the 4 training-test splits described in the previous section. Inference times, intended for a single input, were measured on an i7-6500U CPU averaging the data recorded over the 200 sequences from the test set of the benchmark split of the TCC dataset. The size of the models is the amount of memory occupied on the disk by their serialization.

\begin{table}[t]
    \centering
    \footnotesize
    \begin{tabular}{lcccc}
    \toprule
    \multirow{2}{*}{ \textbf{Model} } & MAE & Time & Model size \\
    & Avg (Std dev) & \textit{(s)} & \textit{(MB)} \\
    \midrule
    TCCNet & 1.98 (0.18) & 0.60 & 72.1 \\
    C-TCCNet & 1.94 (0.18) & 1.84 & 216.5 \\
    C-TCCNet-C4 & 1.79 (0.23) & 3.33 & 298.6 \\
    \midrule
    \textbf{Avg (Std dev)} & 1.90 (0.10) & 1.92 (1.36) & 195.7 (114.7) \\
    \bottomrule
    \end{tabular}
    \caption{Models performance in terms of Mean Angular Error (MAE), over 4 different splits of the TCC dataset, and inference time, using full sequences.}
    \label{tab:resources-full-sequences}
\end{table}

As the table shows, our proposed C-TCCNet-C4 brings the most substantial accuracy gain, with a 10.61\% increase with respect to TCCNet and an 8.38\% increase with respect to C-TCCNet. However, it is considerably more resource-intensive, suggesting that C-TCCNet-C4 should be used for off-line applications, where space and time are less concerning and can be traded in favour of better accuracy (e.g., data preprocessing upstream of computer vision tasks requiring intrinsic colour information). In terms of inference time and model size, TCCNet is by far the top-performing model, and thus the most suitable option in real-time scenarios and for deployment in circumstances where memory is limited. The performance of C-TCCNet lies in the middle of the other two models: it provides a smaller gain in accuracy compared to C-TCCNet-C4 and it is still rather demanding in terms of resources, thus it would be a sub optimal choice for both real-time and off-line applications.

A factor affecting inference time for all the considered models is the number of frames used for illuminant estimation (i.e., the estimating frames), which we discuss in the next section. 

\subsubsection{Impact of number of estimating frames}
\label{sec:impact-seq-len}

\begin{figure*}[t]
    \centering
     \includegraphics[width=.8\textwidth]{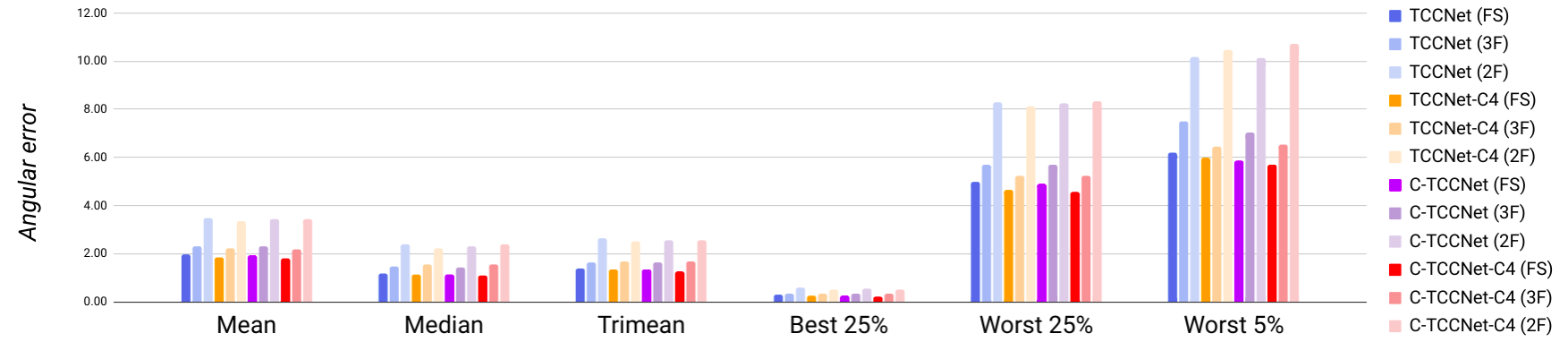}
    \caption{Trends of the angular error on Full Sequences (FS) vs 3 Frames (3F) vs 2 Frames (2F).}
    \label{fig:impact-seq-len}
\end{figure*}

\begin{table*}[t]
    \centering
    \small
    \begin{tabular}{lcc|cc}
    \toprule
    \multirow{2}{*}{\textbf{Model}} & \multicolumn{2}{c}{\textbf{Full sequences}} & \multicolumn{2}{c}{\textbf{3 frames}} \\
    \cmidrule{2-5}
    & MAE & Time \textit{(s)} & MAE & Time \textit{(s)} \\
    \midrule
    TCCNet & 1.98 (0.18) & 0.60 & 2.31 (0.13) & 0.29 \\
    C-TCCNet & 1.94 (0.18) & 1.84 & 2.29 (0.13) & 0.80 \\
    C-TCCNet-C4 & 1.79 (0.23) & 3.33 & 2.20 (0.11) & 1.57 \\
    \midrule
    \textbf{Avg (Std Dev)} & 1.90 (0.10) & 1.92 (1.36) & 2.27 (0.06) & 0.89 (0.65) \\
    \bottomrule
    \end{tabular}
    \caption{Models performance in terms of Mean Angular Error (MAE) and inference time, using full sequences vs 3 frames. The MAE is reported as average over 4 different splits of the TCC dataset and related standard deviation.}
    \label{tab:resources-impact-seq-len}
\end{table*}

Intuitively, a way to reduce inference time for our TCC models is to perform illuminant estimation using shorter sequences of estimating frames, namely use a lower value for the parameter $N$ in Formula \ref{eq:tcc-task}. However, it is reasonable to expect that this approach may impact estimation accuracy, and as a matter of fact \cite{tcc} report preliminary results suggesting that this might be the case for TCCNet. These results involved an informal comparison using full sequences (as we did in Section \ref{sec:accuracy-full-sequences}) vs. sequences of 5 estimating frames, consisting of the shot frame and the four contiguous frames preceding it. Here we provide a more formal analysis of the impact of sequence length on accuracy and efficiency of both TCCNet and our proposed models, and we investigate shorter sequences than in \cite{tcc} to push the boundary toward higher efficiency and monitor how accuracy is affected. Namely, we first trained and tested our models on sequences of two frames only (i.e., the shot frame and the preceding frame), providing the minimum amount of local temporal information to the model. Next, we looked at sequences of three frames, but instead of selecting them to be contiguous to the shot frame, we considered the first frame in the sequence, the median one and the shot frame to ascertain the value of bringing more global temporal information into the process. 

Figure \ref{fig:impact-seq-len} shows how models accuracy varies when sequence length is two or three, as described above, or full, in terms of the angular error for all the metrics introduced in Section \ref{sec:metrics}. This plot confirms that there is a trend of reduced length generating worse accuracy. However, whereas going from three to two frames causes a substantial loss in accuracy (+52\% MEA), the loss is much smaller when going from full sequences to three frames (+19.05\% MEA). This difference is still statistically significant for our best performing model on full sequences, namely C-TCCNet-C4, but it is no longer statistically significant for TCCNet and C-TCCNet\footnote{Accuracy was compared via a one-way ANOVA with post hoc testing performed via Tukey HSD, which adjusts for multiple comparisons.}. In contrast with the limited loss in accuracy, using our selected sequences of length three substantially reduces inference time compared to using full sequences.  Table \ref{tab:resources-impact-seq-len} provides a direct comparison between accuracy in terms of MAE and inference time (computed as described in the previous section) when using full sequences vs sequences of three frames. As the table shows, processing shorter sequences of three frames reduces the inference time considerably (a 53.25\% reduction across models, see last row of Table \ref{tab:resources-impact-seq-len}).

Combined, these results indicate that using three frames, selected as we discussed to capture global temporal information over the full sequences, is a good compromise between accuracy and efficiency, at least for the videos in the TCC dataset (including sequences of 7.3 frames on average with minimum length of 3, maximum of 17, and very few sequences with length below 5 or above 13). 

Of the two models that from our analysis resulted to be most resilient to the reduction of sequence length in terms of accuracy, namely TCCNet and C-TCCNet, the former is the one that achieves the best trade-off between accuracy and efficiency, and thus the most suitable for deployment in real-time scenarios, where inference time is a prominent concern. On the other hand, C-TCCNet-C4 still retains the highest accuracy when trained and tested on full sequences and thus is the model to consider when time and space resources are not an issue. Interestingly, reducing the number of sequences as we did, in general, reduces the differences in accuracy among all models, suggesting that our proposed strategy of cascading TCCNet architectures makes a difference in capturing the richer temporal information available in long sequences.

\section{Conclusions and Future Work}
\label{sec:conclusions}

\noindent In this work, we investigate three novel deep learning architectures for Temporal Colour Constancy (TCC) that build on TCCNet, a neural approach that showed the top accuracy for this task. The first architecture, named TCCNet-C4, attempts to improve TCCNet by substituting its backbone CNN with C4, i.e., the state-of-the-art for single-frame CCC. The other two plug TCCNet and TCCNet-C4, respectively, into a cascading strategy. This cascading strategy resulted in a statistically significant improvement in accuracy over TCCNet when using the full sequences offered by the TCC dataset, which is, at present, the only collection of data specifically designed for CCC in the temporal scenario. 

We performed a novel analysis of the impact of the sequence length and frames selection on accuracy and inference time, finding that, for some of the models in exam, shorter sequences can be processed preserving accuracy while decreasing inference time. Whereas TCCNet proved to be the best option for real-time applications, especially benefiting from the speed boost derived from running on a lower number of frames, our proposed C-TCCNet-C4 provides the top accuracy when time and space are less of a concern. Looking ahead, we plan to further evaluate the performance of the proposed models when using different numbers and selections of frames preceding the one for which the illuminant is estimated. This will allow us to assess the size and configuration of the window of frames that optimizes the trade-off between model accuracy and computational resources needed for training and inference.

With regards to our proposed cascading architectures, we plan to analyze how model performance depends on the number of stages in the cascade, both in terms of accuracy and efficiency. For instance, C-TCCNet-C4 leverages a double cascading strategy (i.e., the outer iteration of the TCCNet-C4 architecture and the inner iteration inside the C4 submodule), which lends itself to be fine-tuned. Furthermore, in these cascading architectures, we applied a very straightforward approach to the iterative colour correction of the sequences, that is correcting the whole sequence based on the estimated illuminant for the shot frame. We plan to look into more sophisticated corrections (e.g., accounting for an estimate of the illuminant for each frame across the sequence). Finally, we want to investigate how to improve the interpretability of our proposed architectures (e.g., using attention-based methods) to better ground the reasons for the detected improvements in performance and to point out possible directions for further refinement.

{\small
\bibliographystyle{ieee_fullname}
\bibliography{paper}

\begin{thebibliography}{10}\itemsep=-1pt

\bibitem{ffcc}
Jonathan~T. Barron and Yun-Ta Tsai.
\newblock Fast fourier color constancy.
\newblock In {\em CVPR}, 2017.

\bibitem{benjamini-hochberg}
Yoav Benjamini and Yosef Hochberg.
\newblock Controlling the false discovery rate: A practical and powerful
  approach to multiple testing.
\newblock {\em Journal of the Royal Statistical Society. Series B
  (Methodological)}, 57(1):289--300, 1995.

\bibitem{quasi-unsupervised-cc}
S. {Bianco} and C. {Cusano}.
\newblock Quasi-unsupervised color constancy.
\newblock In {\em 2019 IEEE/CVF Conference on Computer Vision and Pattern
  Recognition (CVPR)}, pages 12204--12213, 2019.

\bibitem{digital-pathology}
Francesco Bianconi, Jakob~N. Kather, and Constantino~Carlos Reyes-Aldasoro.
\newblock Experimental assessment of color deconvolution and color
  normalization for automated classification of histology images stained with
  hematoxylin and eosin.
\newblock {\em Cancers}, 12(11), 2020.

\bibitem{nus-dataset}
Dongliang Cheng, D. Prasad, and M. Brown.
\newblock Illuminant estimation for color constancy: why spatial-domain methods
  work and the role of the color distribution.
\newblock {\em Journal of the Optical Society of America. A, Optics, image
  science, and vision}, 31 5:1049--58, 2014.

\bibitem{sfu-grayball}
Florian Ciurea and Brian Funt.
\newblock A large image database for color constancy research.
\newblock In {\em Imaging Science and Technology Eleventh Color Imaging
  Conference}, pages 160--164, 01 2003.

\bibitem{imagenet}
J. Deng, W. Dong, R. Socher, L.-J. Li, K. Li, and L. Fei-Fei.
\newblock {ImageNet: A Large-Scale Hierarchical Image Database}.
\newblock In {\em CVPR09}, 2009.

\bibitem{effect-size}
Andy Field.
\newblock {\em Discovering Statistics Using IBM SPSS Statistics}.
\newblock Sage Publications Ltd., 4th edition, 2013.

\bibitem{color-constancy}
David~H. Foster.
\newblock Color constancy.
\newblock {\em Vision Research}, 51(7):674 -- 700, 2011.
\newblock Vision Research 50th Anniversary Issue: Part 1.

\bibitem{color-checker-paper}
P.~V. {Gehler}, C. {Rother}, A. {Blake}, T. {Minka}, and T. {Sharp}.
\newblock Bayesian color constancy revisited.
\newblock In {\em 2008 IEEE Conference on Computer Vision and Pattern
  Recognition}, pages 1--8, 2008.

\bibitem{ccc-survey}
A. {Gijsenij}, T. {Gevers}, and J. {van de Weijer}.
\newblock Computational color constancy: Survey and experiments.
\newblock {\em IEEE Transactions on Image Processing}, 20(9):2475--2489, 2011.

\bibitem{gamut}
Arjan Gijsenij, T. Gevers, and Joost Weijer.
\newblock Generalized gamut mapping using image derivative structures for color
  constancy.
\newblock {\em International Journal of Computer Vision}, 86:127--139, 01 2010.

\bibitem{screening-tools}
Keith Goatman, David Whitwam, A A, John Olson, and Peter Sharp.
\newblock Colour normalisation of retinal images.
\newblock 01 2003.

\bibitem{fc4}
Y. {Hu}, B. {Wang}, and S. {Lin}.
\newblock Fc\textsuperscript{4}: Fully convolutional color constancy with
  confidence-weighted pooling.
\newblock In {\em 2017 IEEE Conference on Computer Vision and Pattern
  Recognition (CVPR)}, pages 330--339, 2017.

\bibitem{squeezenet}
Forrest~N. Iandola, Matthew~W. Moskewicz, Khalid Ashraf, Song Han, William~J.
  Dally, and Kurt Keutzer.
\newblock Squeezenet: Alexnet-level accuracy with 50x fewer parameters and
  {\textless}1mb model size.
\newblock {\em CoRR}, abs/1602.07360, 2016.

\bibitem{alexnet}
Alex Krizhevsky, Ilya Sutskever, and Geoffrey~E. Hinton.
\newblock Imagenet classification with deep convolutional neural networks.
\newblock In {\em Proceedings of the 25th International Conference on Neural
  Information Processing Systems - Volume 1}, NIPS'12, page 1097–1105, Red
  Hook, NY, USA, 2012. Curran Associates Inc.

\bibitem{video-sequences}
V. {Prinet}, D. {Lischinski}, and M. {Werman}.
\newblock Illuminant chromaticity from image sequences.
\newblock In {\em 2013 IEEE International Conference on Computer Vision}, pages
  3320--3327, 2013.

\bibitem{rcc}
Y. {Qian}, K. {Chen}, J. {Nikkanen}, J. {Kämäräinen}, and J. {Matas}.
\newblock Recurrent color constancy.
\newblock In {\em 2017 IEEE International Conference on Computer Vision
  (ICCV)}, pages 5459--5467, 2017.

\bibitem{tcc}
Yanlin Qian, Jani Käpylä, Joni-Kristian Kämäräinen, Samu Koskinen, and
  Jiri Matas.
\newblock A benchmark for temporal color constancy, 2020.

\bibitem{flash}
Y. {Qian}, S. {Yan}, J. {Kämäräinen}, and J. {Matas}.
\newblock Flash lightens gray pixel.
\newblock In {\em 2019 IEEE International Conference on Image Processing
  (ICIP)}, pages 4604--4608, 2019.

\bibitem{cc-reweighting}
J. {Qiu}, H. {Xu}, and Z. {Ye}.
\newblock Color constancy by reweighting image feature maps.
\newblock {\em IEEE Transactions on Image Processing}, 29:5711--5721, 2020.

\bibitem{color-pipeline}
Rajeev Ramanath, Wesley Snyder, YJ Yoo, and Mark Drew.
\newblock Color image processing pipeline.
\newblock {\em Signal Processing Magazine, IEEE}, 22:34 -- 43, 02 2005.

\bibitem{conv-lstm}
Xingjian Shi, Zhourong Chen, Hao Wang, Dit-Yan Yeung, Wai-kin Wong, and
  Wang-chun Woo.
\newblock Convolutional lstm network: A machine learning approach for
  precipitation nowcasting.
\newblock In {\em Proceedings of the 28th International Conference on Neural
  Information Processing Systems - Volume 1}, NIPS'15, page 802–810,
  Cambridge, MA, USA, 2015. MIT Press.

\bibitem{rmsprop}
T. Tieleman and G. Hinton.
\newblock {Lecture 6.5---RmsProp: Divide the gradient by a running average of
  its recent magnitude}.
\newblock COURSERA: Neural Networks for Machine Learning, 2012.

\bibitem{chromatic-adaption}
J. von Kries.
\newblock Chromatic adaption, 1902.

\bibitem{video-based-estimations}
Ning Wang, Brian Funt, Congyan Lang, and De Xu.
\newblock Video-based illumination estimation.
\newblock In Raimondo Schettini, Shoji Tominaga, and Alain Tr{\'e}meau,
  editors, {\em Computational Color Imaging}, pages 188--198, Berlin,
  Heidelberg, 2011. Springer Berlin Heidelberg.

\bibitem{video-surveillance}
Qingyuan Wang, Junbiao Pang, Lei Qin, Shuqiang Jiang, and Qingming Huang.
\newblock Justifying the importance of color cues in object detection: A case
  study on pedestrian.
\newblock {\em The Era of Interactive Media}, 1, 10 2013.

\bibitem{ac-bulb}
J. {Yoo} and J. {Kim}.
\newblock Dichromatic model based temporal color constancy for ac light
  sources.
\newblock In {\em 2019 IEEE/CVF Conference on Computer Vision and Pattern
  Recognition (CVPR)}, pages 12321--12330, 2019.

\bibitem{c4}
Huanglin Yu, K. Chen, K. Wang, Y. Qian, Zhaoxiang Zhang, and Kui Jia.
\newblock Cascading convolutional color constancy.
\newblock In {\em AAAI}, 2020.

\end{thebibliography}
}

\end{document}